\documentclass[10pt,twocolumn,letterpaper]{article}

\usepackage{cvpr}
\usepackage{times}
\usepackage{epsfig}
\usepackage{graphicx}
\usepackage{amsmath}
\usepackage{amssymb}
\usepackage{booktabs}
\usepackage{enumerate}   

\usepackage[breaklinks=true,bookmarks=false]{hyperref}

\cvprfinalcopy 

\def\cvprPaperID{****} 

\begin{document}

\title{Overcoming Catastrophic Forgetting by Soft Parameter Pruning}

\author{Jian Peng$^\dagger$, Jiang Hao$^\dagger$, Zhuo Li$^\dagger$, Enqiang Guo$^\ddagger$, Xiaohong Wan$^\uparrow$, Deng Min$^\dagger$, Qing Zhu$^\sharp$, Haifeng Li$^{\dagger*}$ \\ $^\dagger$School of Geosciences and Info-Physics, Central South University \\ $^\ddagger$ School of Civil Engineering and Transportation, South China University of Technology \\$^\uparrow$Brain and Cognitive Science Research Institute, Beijing Normal University   \\ $^\sharp$ Faculty of Geosciences and Environmental Engineering, Southwest Jiaotong University \\ $^*$Corresponding author: lihaifeng@csu.edu.cn }


\maketitle

\begin{abstract}
   Catastrophic forgetting is a challenge issue in continual learning when a deep neural network forgets the knowledge acquired from the former task after learning on subsequent tasks. 
   However, existing methods try to find the joint distribution of parameters shared with all tasks. This idea can be questionable because this joint distribution may not present when the number of tasks increase. On the other hand, It also leads to "long-term" memory issue when the network capacity is limited since adding tasks will "eat" the network capacity. In this paper, we proposed a Soft Parameters Pruning (SPP) strategy to reach the trade-off between short-term and long-term profit of a learning model by freeing those parameters less contributing to remember former task domain knowledge to learn future tasks, and preserving memories about previous tasks via those parameters effectively encoding knowledge about tasks at the same time. The SPP also measures the importance of parameters by information entropy in a label free manner. The experiments on several tasks shows SPP model achieved the best performance compared with others state-of-the-art methods. Experiment results also indicate that our method is less sensitive to hyper-parameter and better generalization. Our research suggests that a softer strategy,i.e. approximate optimize or sub-optimal solution, will benefit alleviating the dilemma of memory. The source codes are available at https://github.com/lehaifeng/Learning\_by\_memory.

\end{abstract}

\section{Introduction}

Humans can learn consecutive tasks and memorize acquired skills and knowledge throughout the lifetime, such as running, biking and reading. This ability, named continual learning, is also crucial to the development of Artificial General Intelligence. Deep neural networks(DNNs) have achieved remarkable success in various fields \cite{Deng2015ImageNet:,Hannun2014Deep,Simonyan2014Very,Lecun2015Deep}, however, the existing models are unable to handle dynamic tasks and data flow because of catastrophic forgetting, i.e. networks would forget the knowledge learned from previous tasks when training on new datasets \cite{McCloskey1989Catastrophic}.methods to mitigate catastrophic forgetting have been proposed in some literatures. For instance, Rusu et al. \cite{Rusu2016Progressive}, Fernando et al. \cite{Fernando2017Pathnet:} and Lomonaco et al. \cite{Lomonaco2017CORe50:} attempted to restore task-specific structures of model, including some layers or modules, but this would suffer from the limitation of complex selection strategies of genetic algorithms and poor utilization of network capacity. Works \cite{Lopez-Paz2017Gradient,Rebuffi2013icarl:} based on rehearsal strategy reinforce previous memories by replaying experience. 

 An ideal learning system could learn consecutive tasks without increasing memory space, computation cost, as well as transfer knowledge from former tasks to current task. Methods of elastic parameters update\cite{Kirkpatrick2016Overcoming,ritter2018online} could meet these demands by finding the joint distribution of a sequence of tasks. However, it can not restore long-term memory mainly because getting the accurate joint distribution is hard and unnecessary in a long sequence of tasks.

We propose a method to address this problem by getting the approximate solution space satisfying all tasks. It can be achieved through searching the approximate solution space from the approximate solution space corresponding to previous tasks.To achieve it, resistance based on the parameter importance are imposed on the update direction of parameters during learning new tasks.   
 
An appropriate evaluation on parameters is expected to meet the following demands: 1) expression precision, catching some essential parameters; 2) the distribution of values is centralized and  polarized to ensure the parameter space separable; 3) unsupervised. Inspired by the idea of parametric pruning, we propose a method to measure the importance of parameters based on the contrastive magnitude of the change of information entropy with and without parameter pruning. To overcome catastrophic forgetting, exerting a resistance force on updating direction of parameters with large information in the training process, to achieve a balance point between the new task and the old one on the loss surface. The contributions of this paper are as follows: 

\begin{enumerate}[(1)]
    \item We propose SPP strategy. It requres smaller network-capacity for every single task, preserving the memory of previous tasks by constraining fewer parameters; visualization analysis shows that our method can retain important parameters according to tasks adaptively.
    \item We measure the importance of parameters by the variation of information entropy without the need of labels, rather than the variance of loss to the presence or absence of connection between units during a training progress;
    \item Experiment results show our method can effectively overcome catastrophic forgetting and improve overall performance with strong robustness and generalization ability in the case of limited network capability.
\end{enumerate}


\section{Related works}

\textbf{Model prune and knowledge distillation:} Parameter pruning methods \cite{LeCun2015Optimal,Hassibi2014Second} are based on the hypothesis that nonessential parameters have little effect on the model’s error after being erased and thus the key point is to search the optimum parameters that can minimum the interference to error . An effective way to narrow the representational overlap between tasks is to lessen coding parameters of representation in the continual learning model in limited capacity. Knowledge distillation pack the knowledge of complex network into a lightweight target network by the mode of teacher-student, and it also be used to tackle the problem of catastrophic forgetting. 

PackNet \cite{Mallya2017Packnet:} sequentially compress multiple tasks into single model by pruning redundancy parameters to overcome catastrophic forgetting. Dual memory network drew on this idea partially to overcome catastrophic forgetting by an external network. Inspired by the idea of model compression, our method utilizes parameter-importance to set up a soft mask rather than hard pruning based on binary mask, it dose not completely truncate the unimportant parameters, but adaptive adjust to later tasks to some extent, shares part of parameters among multiple tasks and save model capacity compared to hard pruning as well as enjoy lowered performance penalties.

\textbf{Regularization strategies:} Those Methods reduce representational overlap among tasks to overcome catastrophic forgetting by regularization such as weights freezing and weight consolidation. 

Weight Freezing, enlightened by distributed encoding of human brain neurons, tries to avoid overlaps between crucial functional modules of tasks. For instance, Path-Net \cite{Fernando2017Pathnet:} sets up a huge neural network, then fixes specific function module of network to avoid being interfered by later tasks. Progressive Neural Network (PNN) \cite{Rusu2016Progressive} allocates separate networks for each task and performs multitasks by progressive expansion strategy. This kind of methods fix important parameters of a task durably to prevent network from forgetting acquired knowledge. However, those methods suffer network capacity exploding from long-term tasks. 

A classic weight consolidation method \cite{Aljundi2017Memory,Zenke2017Continual} is elastic weight consolidation (EWC) \cite{Kirkpatrick2016Overcoming}. EWC, inspired by the mechanism of synaptic plasticity, updates parameters elastically via determining important parameters. This type of method encode more tasks knowledge with less network capacity and lower computation complexity compared with Path-Net and PNN. The upper bound of tasks EWC can learn is constrained by capacity of network which is determined by the model structure. Since the model structure is invariant during learning process, the increased tasks potentially lead to performance of degenerate of EWC.  


\section{Methodology}

\subsection{Motivation}

\begin{figure}[t]
\begin{center}
   \includegraphics[width=1.0\linewidth]{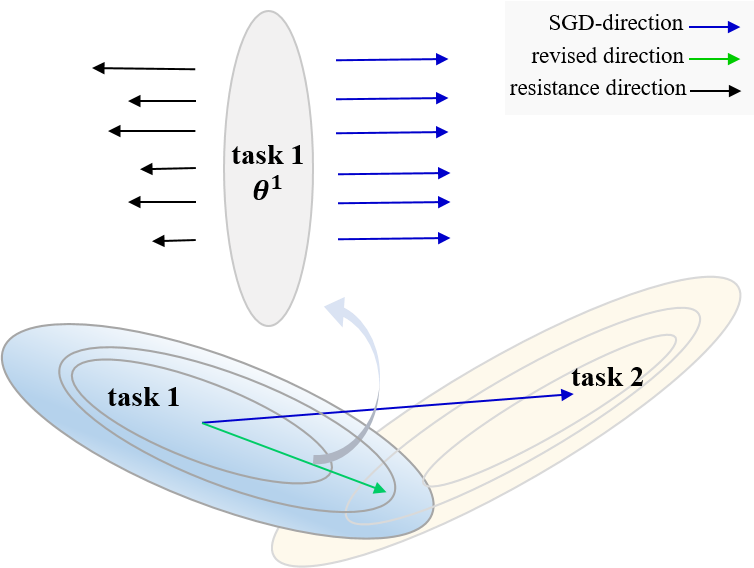}
\end{center}
   \caption{Weight consolidation for overcoming catastrophic forgetting. Blue arrow denotes standard SGD optimizer; black arrow is the resistance about the memories of previous tasks on parameters and green denotes a revised direction of SGD-derection, under the constraint of resistance direction.}
\label{fig:1}
\end{figure}

The cause resulting in catastrophic forgetting is the drift of local minimum point when training a new task.
We claim that it is feasible to approximate a distribution satisfying all tasks through seeking for the current solution space from that of previous tasks. It can be achieved by imposing resistance on parameters in proportional to their importance.(Figure \ref{fig:1}). The key is to ensure the sparseness and representation precision of parameter importance, but there are still some problems in the previous methods of measuring parameter importance: 

\begin{enumerate}[(1)]
    \item Getting importance value of parameters by calculating gradient descent of loss function would unavoidably underestimates these importance value as the model reaches its convergence;
    \item Supervised, it relies heavily on labeled training data and testing data;
    \item The distribution of important parameters is relatively divergent. Calculating the importance of parameters depends on sensitive of a model responding to the parameters perturbation instead of the magnitude of parameters’ weights. Lower magnitude could possibly result in higher model sensitivity than higher ones , without considering the effects of cumulative changes in parameters, resulting in a high occupancy of capacity for each task. 
\end{enumerate}
We design a method to measure the  parameter importance satisfying the concentrated and polarized distribution, and then propose a framework to overcome catastrophic forgetting by SPP.



\subsection{Framework}
Following 3.1, we present the framework of SPP strategy in Figure \ref{fig:2}. We calculate the the coefficient of resistance of parameters on previous $T-1$ tasks after learning the$ (T-1)^{th}$ task. Then when the $T^{th}$ task is coming, we update the direction of gradients according to the former coefficient of resistance. 



\subsubsection{Measure of importance of parameter}

\textbf{Definition}: Given a well-trained model, we try to train parameters W on input X to reduce the error $E=\sum_{i=1}^C p_i \log q_i$, and the model learned can be expressed as $F(X,W)\to E$. If we set $W_k$ as 0，the change of error $\delta E$ corresponding to $W_k$ can be written as $F(X,W,0)-F(X,W,W_k) \to \delta E$. The larger $\delta E$ means the more important $W_k$.
The formula of Taylor expansion is:

\begin{figure}[t]
\begin{center}
   \includegraphics[width=1.0\linewidth]{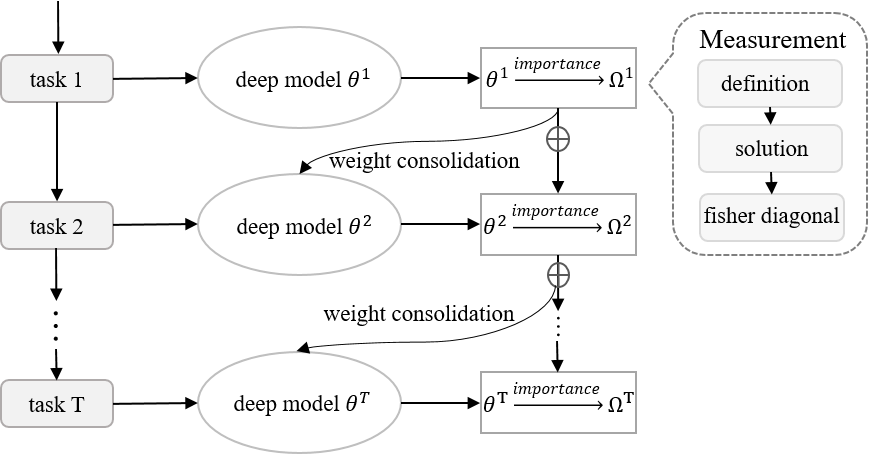}
\end{center}
   \caption{Framework for continual learning on T sequential tasks.}
\label{fig:2}
\end{figure}

\begin{equation}\label{Acer}
    \delta E = (\frac{\partial E}{\partial W})^T\delta W + \frac{1}{2}\delta W^T H\delta W + O(\lVert \delta W\rVert^3)
\end{equation}

$H \equiv \partial ^2 E/\partial W^2$ is the Hessian matrix on parameters; $ \partial E/\partial W$ represents the gradient on the $W$. The gradient will close to 0 when the model converges and the first item on the right side will be too minimal to calculate a precise value of the error change to the parameter perturbation. Therefore second-order approximate solution is used as a satisfying amendment.

\textbf{Solution} - The method above depends on the true distribution p which is calculated by labeled data. To get rid of the need of labels, we use the information entropy approximate the error $E$ because the distribution p and predicted distribution q are proximate on a well-trained model. Thus, the parameter importance is in term of: 

\begin{equation}\label{Acer}
    \delta E = (\frac{\partial H(q)}{\partial W})^T\delta W + \frac{1}{2}\delta W^T H\delta W + O(\lVert \delta W\rVert^3)
\end{equation}

Where $H(q)=\sum_{i=1}^C q_i \log q_i$. The idea behind is to measure the steady state of a learning system utilizing information entropy. We explain it that: the output distribution of model will gradually involve from a random state into a stable state and with entropy decreasing. When the model converges, the system will perform stably on the training data, which means low entropy and certain output distribution. And the statement  of parameters connection would influence the stability, which means the change of entropy.

\textbf{Hessian diagonalization} - The calculation of Hessian is complex and high computation. We introduce the Diagonal Fisher Information Matrix \cite{Pascanu2013Revisiting} to approximate Hessian matrix. The advantage is that its computational complexity is linear and can be solved quickly through gradient. However, the diagonalization may lead to a loss of precision. We speculate that better results can be obtained if we adopt a better Hessian approximation method.

\subsubsection{Cumulative importance computation} Given a set of t+1tasks, we calculate the importance $\Omega_{i,j}^t$ of the parameter $w_{i,j}$ after learning the $t^{th}$ task. Where  i and j represent the connection between the $i^{th}$ neuron and the $j^{th}$ neuron in neural networks respectively. 
\begin{equation}\label{Acer}
    \Omega _{i,j}^t=max(0,\Omega _{i,j}^t)
\end{equation}


According to equation (2), the positive values of parameter importance indicate these parameters are significant to current task and vice versa. Thus, We set the negative value of importance all to 0 to reduce the resistance of learning new tasks. After learning the t+1 task, we accumulate the importance of the previous tasks to obtain the accumulated importance on the parameters on t+1 tasks:

\begin{equation}\label{Acer}
    \Omega_{i,j}^{1:t+1} = \Omega_{i,j}^{1:t} + \Omega_{i,j}^{t+1}
\end{equation}

\subsubsection{Weight consolidation}

To avoid previous memories forgetting, we protect important parameters from being destroyed in subsequent training process by adding additional regular terms in the target function:

\begin{equation}\label{Acer}
    L = L_{new} + \lambda \sum_{i,j}^{W} \Omega_{i,j}^{1:t}(w_{i,j}-w_{i,j}^{'})^{2}  
\end{equation}
$L_{new}$ is the loss function of the current task. Here, $w_{i,j}^{'}$ is the parameter of the model after learning last task, $w_{i,j}$ denotes the parameter corresponding to current task, and $\Omega_{i,j}^{1:t+1}$ denotes the cumulative importance of a parameter in the previous t tasks.And we present roughly our algorithm in Table \ref{tab:1}.

\begin{table}
	\begin{center}
		\caption{Pseudo-code for overcoming catastrophic by soft parameter pruning}
		\label{tab:1}
		\scriptsize
		\begin{tabular}{l}
			\toprule
			\textbf{Overcome Catastrophic Forgetting by Soft Parameter Pruning:}\\
			\midrule
			\textbf{Start with:}\\
			\quad $W_{i,j}^{*}$: old task parameters\\
			\quad $W_{i,j}$: new task parameters\\
			\quad $X,Y$: training data and ground truth on the new task\\
			\quad tasks: all tasks numbers\\
			\quad $H(q)$: information entropy of output\\
			\quad H: Hessian Matrix\\
			\quad $e_{k}^{T}$: the unit vector with correspongding to $W_{k}$\\
			\textbf{Training:}\\
			\quad \textbf{for task$\in$ tasks do}\\
			\qquad $W_{i,j}^{*}$.assign($W_{i,j}$) \qquad //Update old task parameters\\
			\qquad //Calculate the importance of the parameters of the T-1 tasks\\
			\qquad $\Omega_{i,j}^{t}=max(0,(\frac{\partial H(q)}{\partial W_{i,j}})^T\delta W_{i,j}+\frac{1}{2}\delta W_{i,j}^T H\delta W_{i,j})$\\
			\qquad $s.t.e_k^T\delta W_{i,j} + W_k =0$\\
			\qquad$\Omega_{i,j}^{1:t}=\Omega_{i,j}^{1:t-1}+\Omega_{i,j}^t$ \qquad //Cumulative importance computation\\
			\qquad \textbf{Define:} $\hat{Y}=CNN(X,W_{i,j})$ \qquad //new task output\\
			\qquad $W_{i,j}\leftarrow arg_{W_{i,j}}min(L_{new}(Y,\hat{Y})+\lambda  \sum_{i,j}^{W}\Omega_{i,j}^{1:t}(W_{i,j}-W_{i,j}^{*})^2)$\\
			\qquad \qquad //Update new task parameters\\
			\quad \textbf{end for}\\
			\bottomrule
		\end{tabular}
	\end{center}
\end{table}

\section{Experiment and analysis}

\subsection{Experiments setting}

\textbf{Data}. The permuted MNIST \cite{Srivastava2014Compete} or Split MNIST \cite{Lee2015Overcoming} is too simple to evaluate our methods. In order to verify the ability of generalization, we tested the proposed method on three tasks: image classification task with CNN model, long-term incremental learning and generative task with Variational AutoEncoder (VAE) model. 

In image classification task, the Cifar10 \cite{Krizhevsky2009Learning}, the NOT-MNIST \cite{Bulatov2011Notmnist}, the SVHN \cite{Netzer1989Reading}, and the STL-10 \cite{Coates2015An} which are all RGB images with the same size of 32*32 pixels, are chosen. In long-term incremental learning task, Cifar100 \cite{Krizhevsky2009Learning} is used for medium scale network model, and Caltech101 \cite{Fei-Fei2006One-shot} is used for large scale network model (shown in supplement). In the generative task, celebA \cite{Liu2018Large-scale} and anime face crawled from web are selected as test data. These two database share the same resolution. 

\textbf{Baseline} – We compared our method with state of the art methods, including LWF \cite{Li2017Learning}, EWC \cite{Kirkpatrick2016Overcoming}, SI \cite{Zenke2017Continual} and MAS \cite{Aljundi2017Memory}, as well as some classic methods, including standard SGD with single output layer (single-head SGD), SGD with multi-output layers, SGD with the intermediate layers frozen(SGD-F), and fine-tuning intermediate layer(finetuning). We defined a multi-tasks joint training with SGD (Joint) \cite{yuan2012visual} as the baseline to evaluate the difficulty of a sequential tasks.     

\textbf{Evaluation} – We utilize Average Accuracy(ACC), Forward Transfer (FWT), and Backward Transfer (BWT) \cite{Lopez-Paz2017Gradient} to estimate the model performance: (1) ACC, evaluating the average performance of processing tasks; (2) FWT, describing the suppress of former tasks on later tasks; (3) BWT, describing the forgetting of previous tasks. Evaluating the difficulty of individual task through testing the model by multi-tasks joint training \cite{yuan2012visual} is more objective than testing the model of single task. Therefore we put forward a modified version. Given $T$ tasks, we evaluate previous t tasks after trained on the $t^{th}$ task. Denoting the result of i task tested on the $j^{th}$ task model as $P_{j,i}$. We use three indicators:

\begin{equation}\label{Acer}
    ACC(i) =\frac{1}{T} \sum_{i=1}^{T}P_{T,i}  
\end{equation}

\begin{equation}\label{Acer}
    FWT =\frac{1}{T-1} \sum_{i=1}^{T-1}P_{i,i}-m_{i}  
\end{equation}

\begin{equation}\label{Acer}
    BWT =\frac{1}{T-1} \sum_{i=1}^{T-1}P_{T,i}-P_{i,i}  
\end{equation}

Higher value of ACC indicates better overall performance, and higher value of BWT and FWT indicate better trade-off between memorizing previous tasks and learning new ones. 

\textbf{Training} – All models share the same network structure with dropout layer\cite{Goodfellow2013An}, and we initialized all parameters on MLP with random Gaussian distribution which has the same mean and variance ($\mu=0, \sigma=0.1 $ ), and applied Xavier on CNN. We optimized models by SGD with initial learning rate searching from ${0.1, 0.01，0.001}$ with a decay ratio of 0.96, and with uniform batch size. We trained models with fixed epoch and global hyper-parameters for all tasks. We chose the optimal hyper-parameters by greedy search.

\subsection {Experiment results and analysis}
\subsubsection{MLP\& MNIST}

\textbf{Split MNIST}

We divided the data into 5 sub-datasests, and trained an MLP with 784-512-256-10 units. In Table \ref{tab:2}, we present experiment results with split MNIST. Not all continual learning strategies work well on all indexes. Fine-tuning and SGD perform best on FWT, because no free memories are required for the subsequent tasks, and some features may be reused to improve learning of the new tasks while tasks are similar. LWF, MAS and SI perform well on BWT and ACC, and our method achieve the best performance on both indexes besides the joint learning method. We conclude that the model learns the general features from multiple datasets, which means models implicitly benefit from data augmentation. Our results of ACC and FWT can rival the best ones in single index and our model has the least catastrophe forgetting problem on BWT and it has only a 1.5 reduction in ACC after learning 10 tasks. In general, our method outperforms another eight approaches.

\begin{table}
\begin{center}
\caption{The results of Split-MNIST}
\label{tab:2}
\begin{tabular}{lccl}
\toprule
Method & FWT(\%) & BWT(\%) & ACC(\%) \\
\midrule
SGD & -0.31 & -34.01 & 61.53\\
SGD-F & -18.6 & -12.9 & 84.82\\
Fine-tuning & \textbf{-0.29} & -13.9 & 82.04\\
EWC \cite{Kirkpatrick2016Overcoming} & -4.99 & -6.43 & 88.75\\
SI \cite{Zenke2017Continual} & -6.19 & -3.51 & 90.67\\
MAS \cite{Aljundi2017Memory} & -4.38 & -2.08 & 94.09\\
LWF \cite{Li2017Learning} & -4.42 & -2.04 & 94.08\\
Joint \cite{yuan2012visual} & / & / & \textbf{99.87}\\
Ours & -0.44 & \textbf{-0.75} & 98.31\\
\bottomrule
\end{tabular}
\end{center}
\end{table}

\textbf{Permuted MNIST}
We evaluate our method on 10 permuted MNIST tasks. In Table \ref{tab:3}, we present the results of our approaches and those of others. As what we expected, our method performs best on FWT, which are superior to SGD, we own it to the possibility of some features of lower layer can be shared by new task and there is enough capacity to relieve the pressure on capacity demand in new tasks. SGD-F gets the highest score on BWT, because SGD-F has fixed parameters, which help protecting the parameters of previous tasks from being overwritten, but it is at cost of the disability to learn new tasks flexibly. LWF performs worst on permuted MNIST compared with the split MNIST despite a good score, which may be attributed to dataset changing mentioned above on FWT. And our method gets the comparable performance on ACC.    

\begin{table}
\begin{center}
\caption{The results of permuted MNIST of 10 tasks}
\label{tab:3}
\begin{tabular}{lccl}
\toprule
Method & FWT(\%) & BWT(\%) & ACC(\%) \\
\midrule
SGD & 1.11 & -18.05 & 70.45\\
SGD-F & -14.90 & \textbf{0.10} & 81.99\\
Fine-tuning & 0.75 & -6.21 & 80.69\\
EWC \cite{Kirkpatrick2016Overcoming} & -0.98 & -2.57 & 91.97\\
SI \cite{Zenke2017Continual} & -0.56 & -4.40 & 90.21\\
MAS \cite{Aljundi2017Memory} & -1.23 & -1.61 & 92.6\\
LWF \cite{Li2017Learning} & 0.67 & -24.02 & 74.15\\
Joint \cite{yuan2012visual} & / & / & \textbf{95.05}\\
Ours & \textbf{2.33} & -3.22 & 94.51\\
\bottomrule
\end{tabular}
\end{center}
\end{table}


\subsubsection{CNN \& image recognition}
\textbf{Sequence of image recognition tasks}

Further, we test out method on nature visual datasets based on VGG \cite{Simonyan2014Very} with 9 layers and batch normalization layer to prevent gradient exploding. Specifically, we train and test on MNIST, notMNIST, SVHN, STL-10 and Cifar10 in this order, which have been processed to the same amount of train images and categories (50,000 and 10, respectively). 

\begin{figure}[t]
\begin{center}
   \includegraphics[width=1.0\linewidth]{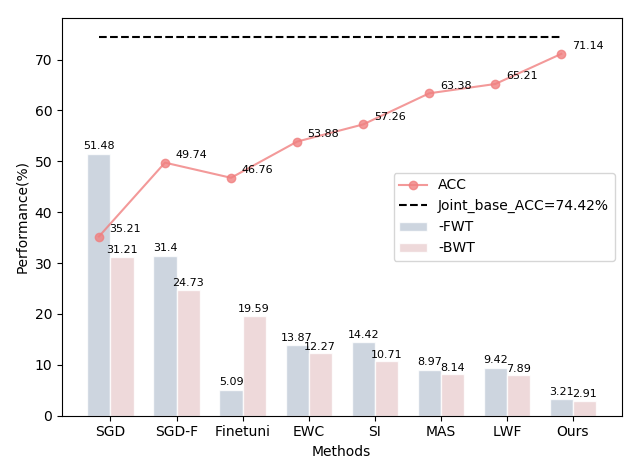}
\end{center}
   \caption{The performance of different overcoming catastrophic forgetting methods on the sequence of visual datasets. The method based on regularization produces a little effect starting from EWC, although the effect is limited; MAS and LWF are close; Our method reaches the best performance on all the indicators.}
\label{fig:3}
\end{figure}

Overall, our method achieved the best performance on FWT, BWT and ACC. As Figure \ref{fig:3} shown, our method is almost one-third of LWF and MAS on FWT. It indicates that our proposed method work well on alleviating the dilemma of memory, and the test accuracy is close to the baseline nearly. Namely, our method drives the network to train well on the sequence of tasks and on BWT, ours also reaches the top result, which means that it ensures the network still have a good ability to handle the old tasks after the continual training. On ACC, our method has also achieved nearly performance as multi-task joint training, which shows that networks can do a good trade-off among tasks. The result of Fine-tuning is better than that of SGD, indicating that the configuration of independent classifier for each task can prevent catastrophic forgetting to some extent. We speculate that it is because the features of different tasks at the high layer are highly entangled, and using different classifiers can alleviate this situation a little.

\subsubsection{Robust analysis}

\begin{figure}[t]
\begin{center}
   \includegraphics[width=1.0\linewidth]{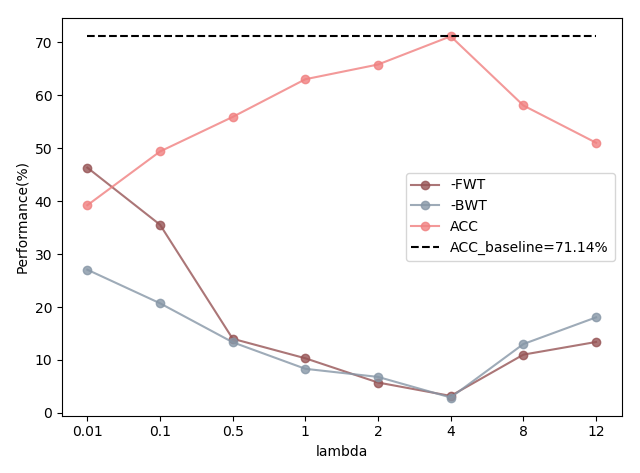}
\end{center}
   \caption{The performance of SPP under different hyper-parameter. The horizontal axis represents the value of hyper-parameter. The vertical axis represents the result of three indicators. The dotted black line indicates the baseline of accuracy.}
\label{fig:4}
\end{figure}

To test the stability of our method to hyper-parameter, based on the above experiment, we test the method under different $\lambda$. The result shows that ours have a strong robustness to hyper-parameter in a large range of values and can overcome strophic forgetting to some extent. As shown in Figure \ref{fig:4}, when $\lambda$ is 0.01, the network only focuses on the training on the new task and does not care about the protection of the old task. In this case, all the three indicators are extremely poor, and the proposed method and SGD are almost the same at this time. When $\lambda$ reaches 0.1, the proposed method has achieved relatively good performance and has greatly improved on all three indicators. If $\lambda$ is in the range of 0.5 to 4,the performance is relatively stable. The proposed method achieves the best performance with $\lambda=4$. As $\lambda$ continues rise, the network memorizing too much results in lack of capacity to learn, which makes the performance of new task processing decreased. 

\subsubsection {Continual learning in VAE}	 

To test the generalization of our method, we apply it in VAE, we carry out tasks in sequence from human face to anime face. We resize the samples of two datasets to the same size of 96*96, and train a VAE with conv-conv-fc encoder layer and fc-deconv-deconv layer on both sides. Then we use separate latent variable to train single task, which is essential for the performance of VAE because of significant difference of distributions between two tasks.

We trained models by three manners: (1)training on the Celeba dataset from scratch; (2) training on the Celeba and then training on the anime face with SGD; (3) training on the Celeba and then training on the anime face with SPP.

\begin{figure*}[t]
\begin{center}
   \includegraphics[width=1.0\linewidth]{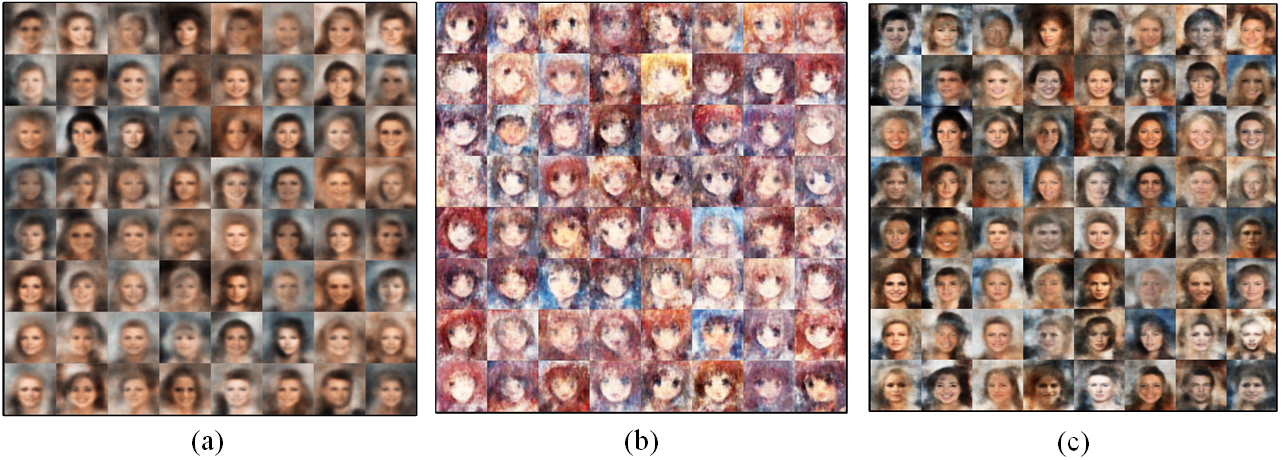}
\end{center}
   \caption{Overcome catastrophic forgetting from face dataset to anime dataset use VAE. To guarantee the objectivity of the results, we respectively utilize different data and different network structures. left: The test sample of human face with generator from the human face dataset; middle: The test sample of human face with generator after trained from celebA to anime face dataset, without using our approach; right: the test sample of human face with generator after trained from celebA to anime face dataset using our approach.}
\label{fig:6}
\end{figure*}

In Figure \ref{fig:6}, we present the samples of human face produced by three models. The results show that our approach can well preserve the skill of human face generation while learning anime face. The model with SPP works well as the model train on the Celeba, but the model with SGD loses the ability. it p roves that SPP has strong generalization.   

\subsubsection{Discussions}
\begin{figure*}[t]
\begin{center}
   \includegraphics[width=1.0\linewidth]{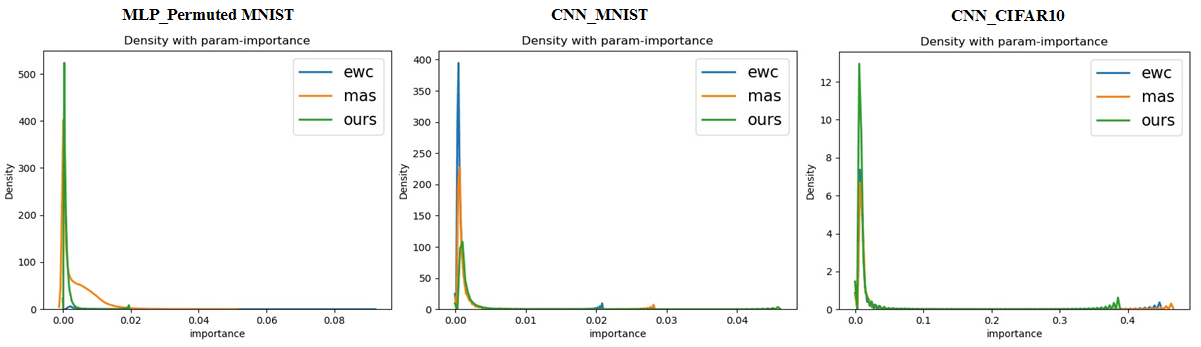}
\end{center}
   \caption{Distribution of parameter-importance. The horizontal axis is importance value; the vertical axis is density; the blue solid line is the calculation result of Fisher informationmatrix in EWC; the orange solid line represents the method of MAS; the green solid line is the result of our method. Left: the distribution of parameter-importance measure on MLP model trained with Permuted-MNIST; mid: the result measure on vgg9 trained with MNIST; right: the result measure on vgg9 trained with CIFAR10}
\label{fig:7}
\end{figure*}

\textbf{Analysis of parameter-importance.} As mentioned above, we expect the distribution of parameter-importance is concentrated and polarized. In Figure \ref{fig:7}, We show the distribution of parameter importance obtained by the three methods. The results shows that a distribution with these two characters contributes to overcoming catastrophic forgetting. The left figure shows the our distribution is sharp at low importance and high importance, indicating that our method frees more parameters to learn more tasks. And the figure on the right shows similar results based on CNN and CIfar10. Compared with other methods, the distance between peaks of the distribution is closer shown in the middle figure. We speculate the absolute distance at is large enough(0-0.4) to distinguish different tasks. As shown in the right figure, the overall parameter importance is low in values, we believe that the capacity of vgg9 for MNIST is sufficient enough, and compared with other methods, our method is also polarized.

\begin{figure*}[t]
\begin{center}
   \includegraphics[width=1.0\linewidth]{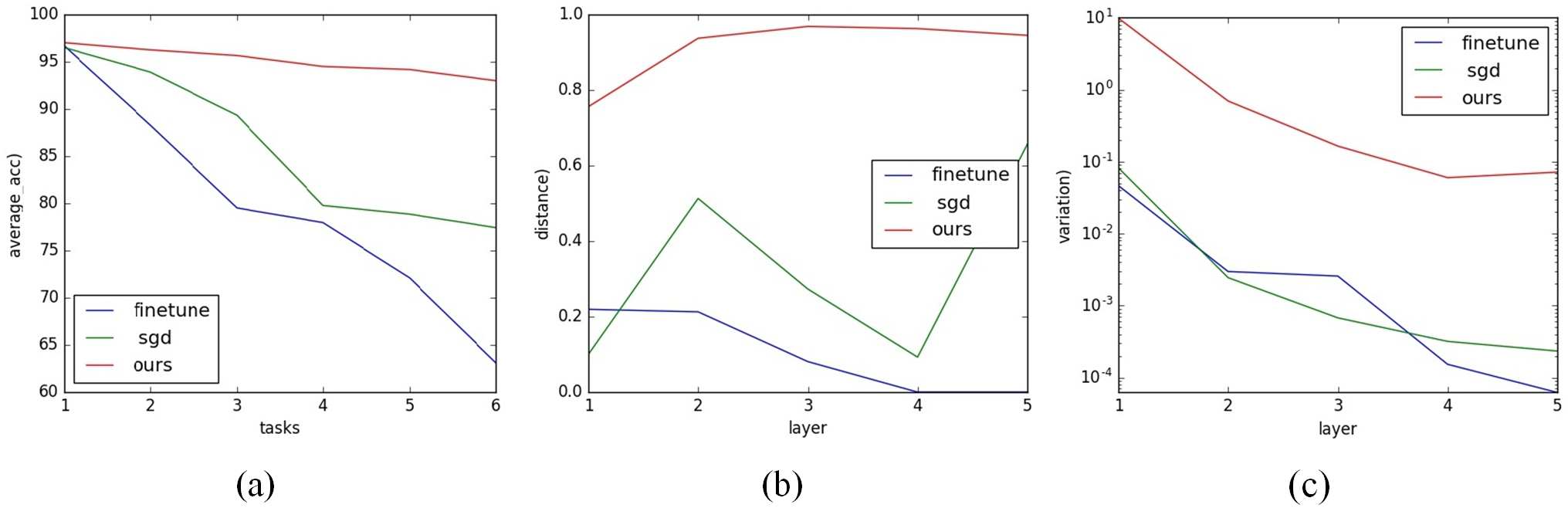}
\end{center}
   \caption{Parameter space similarity and changing analysis on Pemuted-MNIST sequential tasks, a red line denotes our method, a blue line denotes fine-tuning and a green line denotes standard SGD with single head; (a): Overall average accuracy in 6 permuted MNIST sequential sub tasks; (b): Similarly of parameter space; (c): Parameter variance between parameters of tasks.}
\label{fig:8}
\end{figure*}

\textbf{Parameter space similarity and changing analysis.} We carried out six sequential tasks using our method with Permuted-MNIST and analyze the experiment results in comparison with SGD of single-head and Fine-tuning of multi-head as control groups: 
\begin{enumerate}[(1)]
    \item The evolving of overall average accuracy is shown in Figure \ref{fig:8}(a) which indicates that our method is more stable and achieves better results as growing number of tasks;
    \item In order to verify model can preserve previous memory efficiently, we use Fréchet distance \cite{frechet1906quelques} to measure the similarity of parameter importance distribution between the first tasks and the last tasks, Figure \ref{fig:8}(b). In general, F value in our method is far greater than other two methods, which indicates that our method can better retain the important parameters of previous task, and the F values are greater in deeper layers of networks, which reveals that strengthening protection of parameters on deep layers may tremendously help tackle catastrophic forgetting;
    \item In Figure \ref{fig:8}(c), We utilized weighted sum of squares of difference between the first and the last task to measure the parameters' change. Our results show that parameters in deeper layers  change less, and the fluctuation of parameters based on our methods is much larger than other methods. It indicates our method can preserve the former memories but it leads to larger network capacity to learn new tasks.
\end{enumerate}

\begin{figure}[t]
\begin{center}
   \includegraphics[width=1.0\linewidth]{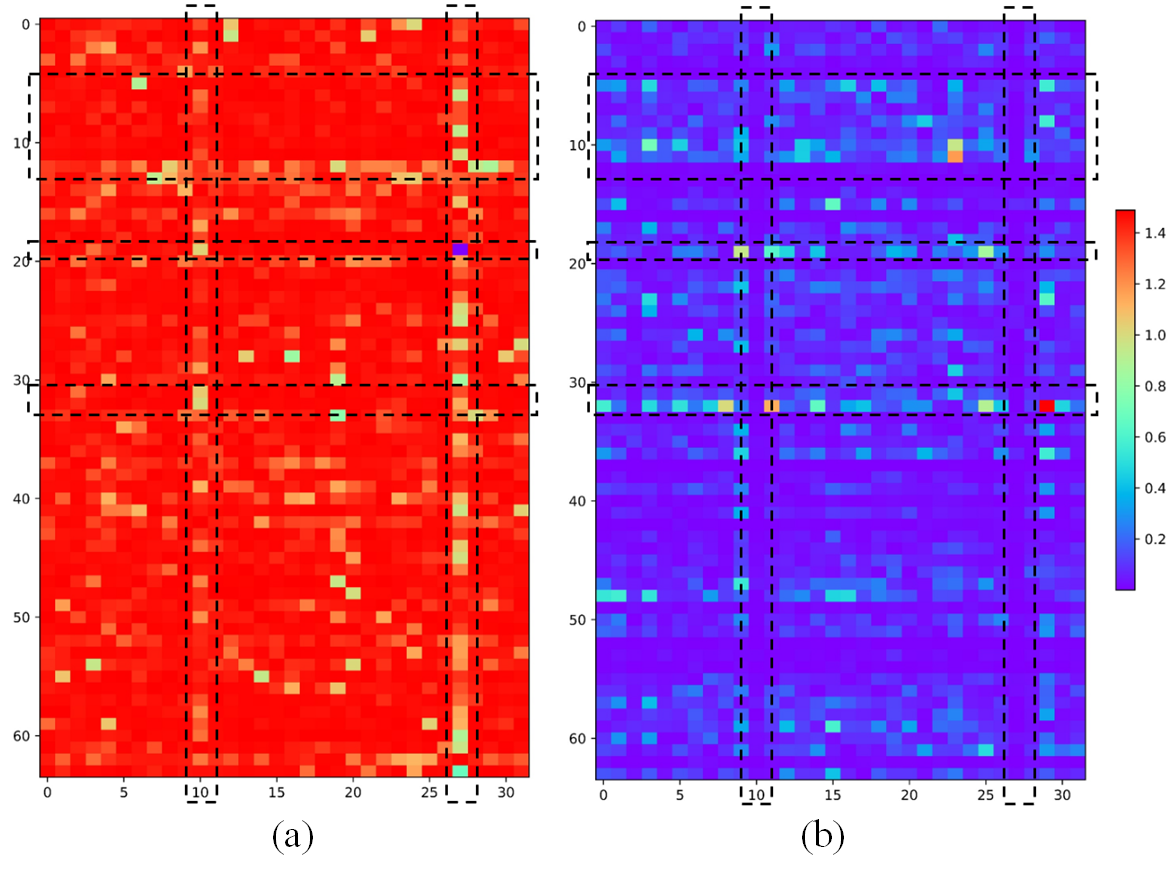}
\end{center}
   \caption{Visualization of importance and variance of parameters. The horizontal axis represents the neurons of the output layer, the vertical axis represents the neurons of the input layer, and each element represents the connection between the neurons of the input and output layer. Left: variance of parameters between two tasks, the colder the color is, the smaller the variance is; right: importance of parameters of the first task, the warmer the color is, the more significant the parameter is.}
\label{fig:9}
\end{figure}

\textbf{Visualization analysis.} We visualize the negative of absolute value of parameters change(left), and compare it with the distribution of parameter importance(right). Our results show that our method can prevent significant parameters  from being updated and make full use of non-significant parameters to learn new tasks. In Figure \ref{fig:9}, in the black dotted bordered rectangle of the $1^{st}$ row, parameters with warm color change little. In contrast, parameters in the second column of the picture on right are unimportant and they change hugely. It indicates that our method can precisely capture the significant parameters and prevent them from being updated to prevent forgetting.

\section{Conclusion and future works}
In this paper, we proposed a Soft Parameters Pruning (SPP) strategy to overcome catastrophic forgetting by find a trade-off between short-term and long-term profit of a learning model. Our strategy tries to free those parameters less contributing to remember former task domain knowledge to learn future tasks, and preserve memories about previous tasks via those parameters effectively encoding knowledge about tasks at the same time. The SPP strategy also catches parameters with high information and prevent them from being overwritten in a soft way to prevent forgetting. Experiments show some advantages of SPP strategy: 
\begin{enumerate}[(1)]
    \item Defining a measurement strategy guaranteeing the precision; 
    \item Our approach is low-sensitive to hyper parameters;
    \item Our approach can be extended to generative model.
\end{enumerate}

Our evidences suggest that that finding a approximate optimize or sub-optimal solution will benefit alleviating the dilemma of memory. We also find that concentration and polarization properties of parameters distribution are significant for overcoming catastrophic forgetting.

The aim of overcoming forgetting in long-sequence tasks has not been fully achieved because of protecting some parameters through measurement based on single strategy is not entirely convincing. We suggest that well-structured constraints to control parameters behavior or well-designed pattern of parameters distribution may be crucial to the good performance of a model to overcome forgetting. Also, research on human brain memory is considering a potential way to solve this problem \cite{Hassabis2017Neuro}. The problem of overcoming catastrophic forgetting is still open.

 


{\small
\bibliographystyle{ieee}
\bibliography{egbib}
}

\title{Supplementary Materials}
\maketitle
\section{Incremental learning}

\textbf{Large scale dataset from Caltech-101} To test the performance of our method on a larger dataset, we randomly splitted the Caltech-101 into 4 subsetswith 30,25,25,22 classes respectively, and then divided each part of them into training and validation set according to the ratio of 7:3. In the experiment, we resized the images to [224,224,3], normalized the pixels into [0,1] and randomly flipped the images left and right to augment the data in preprocess. We employed ResNet18 as the basic network. Because the categories of four datasets are not consistent, we added a new separate classifier and a fully connected layer before the classifier for each task. Each new fc layer has 2048 neural units and the dropout rate is set to 0.5. The iteration size and batch size of every task are 100 epochs and 128. The initial learning rate is set to 0.001, using decay with every 100 epochs to 90\% of the original. In order to prevent overfitting, we search the hyper parameter randomly ranging from 0.5 to 30. Due to the inconsistent numbers of categories in four subsets, we do not compare our method with SI.  

A well-functioning model is expected to be stable under abrupt change of tasks. In order to evaluate the stability of model on unseen tasks, we designed an indicator SMT as follow:

\begin{equation}\label{Acer}
    SMT =\frac{1}{T-1} \sum_{j=1}^{T-1}D_{j}  
\end{equation}

Where $D$ is the variance of one task for sequential learning, which reflects the performance fluctuations of a task. 

\begin{figure*}[h]
\begin{center}
   \includegraphics[width=1.0\linewidth]{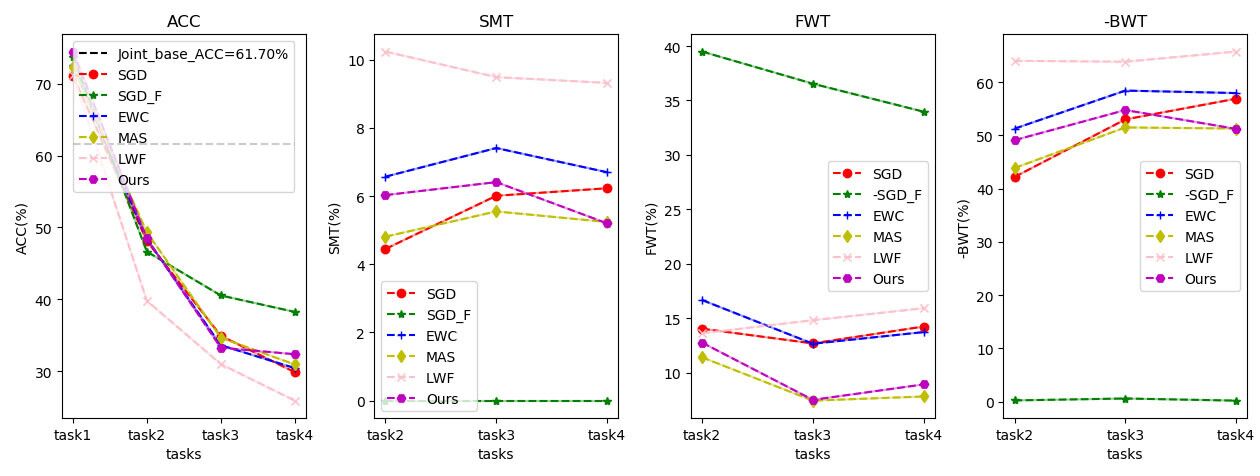}
\end{center}
   \caption{Performance for subset of Caltech101. The x-axis denotes the tasks trained on resnet-18 and y-axis denotes the indicator of ACC, SMT, FWT and BWT, noting that we presented the negative value of the FWT and BWT in figures.}
\label{fig:11}
\end{figure*}

\textbf{Long sequence for CIFAR100.} As shown in Figure \ref{fig:11}, all current methods do not perform well in large scale datasets as the number of tasks increasing. In the fourth task, the ACC of our method is less than that of SGD-F. However but outperform EWC, MAS and LWF. In SMT, when the model learns the second and third task, our method is worse than SGD and MAS. When it comes to the fourth task, our method works better than all the rest methods in BWT and SMT, which shows that our method can keep the good memory of tasks with longer sequence and was better stability. In FWT, our method shows the best performance except the MAS. Overall, our method is better than state-of-the-art methods based on regularization.   

\begin{figure*}[h]
\begin{center}
   \includegraphics[width=1.0\linewidth]{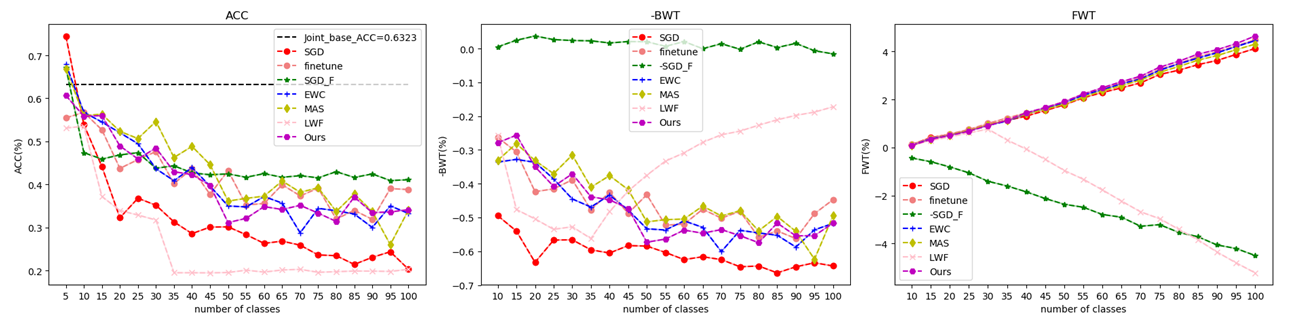}
\end{center}
   \caption{Performance for incremental learning on CIFAR100. The x-axis denotes the tasks trained on vgg with 9 layers, each task contains 5 categories; the y-axis denotes the indicator of ACC, FWT and BWT.}
\label{fig:12}
\end{figure*}

The results in Figure \ref{fig:12} indicate that it is still difficult to make models capable of long-term memory, especially in complex tasks. Our method achieved similar results in the overall performance among regularization methods, but SGD-F and finetuning performed better when the number of learning task is large, and LWF almost lost its learning ability. On BWT, our method and MAS achieved a better result. SGD-F performed best on preventing forgotten, because the weights were completely fixed. Noting that LWF shows higher BWT, but the data is useless due to the loss of learning ability. On FWT, our method achieved the best results, which indicating that our method has little impact on the learning of new tasks while preserving previous knowledge.

\end{document}